\documentclass[a4paper,fleqn]{cas-dc}

\usepackage[authoryear,longnamesfirst]{natbib}
\usepackage{threeparttable}
\usepackage{natbib}
\usepackage{subfigure}
\usepackage{caption}
\usepackage{multirow}
\usepackage{amsmath,amssymb}
\usepackage{algorithmic}
\usepackage{graphicx}
\usepackage{amsmath,amssymb}
\usepackage{textcomp}
\usepackage{arydshln}
\usepackage{subfigure}
\usepackage{bbding}
\usepackage{utfsym}

\newcommand{\etal}{\textit{et al}.}
\def\tsc#1{\csdef{#1}{\textsc{\lowercase{#1}}\xspace}}
\tsc{WGM}
\tsc{QE}

\begin{document}
\let\WriteBookmarks\relax
\def\floatpagepagefraction{1}
\def\textpagefraction{.001}



\title [mode = title]{CWP: Instance complexity weighted channel-wise soft masks for network pruning} 



%
 \author[1]{Jiapeng Wang}[type=editor, style=chinese]
 \author[1]{Ming Ma}[type=editor, style=chinese]
 \author[1,2]{Zhenhua Yu}[type=editor, style=chinese]
 \cormark[1]
 \cortext[cor1]{Corresponding author}
 \ead{zhyu@nxu.edu.cn}
 \address[1]{School of Information Engineering, Ningxia University, Yinchuan, 750021, China}
 \address[2]{Collaborative Innovation Center for Ningxia Big Data and Artificial Intelligence co-founded by Ningxia Municipality and Ministry of Education, Yinchuan, 750021, China}









\begin{abstract}
Existing differentiable channel pruning methods often attach scaling factors or masks behind channels to prune filters with less importance, and implicitly assume uniform contribution of input samples to filter importance. Specifically, the effects of instance complexity on pruning performance are not yet fully investigated in static network pruning. In this paper, we propose a simple yet effective differentiable network pruning method CWP based on instance complexity weighted filter importance scores. We define instance complexity related weight for each instance by giving higher weights to hard instances, and measure the weighted sum of instance-specific soft masks to model non-uniform contribution of different inputs, which encourages hard instances to dominate the pruning process and the model performance to be well preserved. In addition, we introduce a regularizer to maximize polarization of the masks, such that a sweet spot can be easily found to identify the filters to be pruned. Performance evaluations on various network architectures and datasets demonstrate CWP has advantages over the state-of-the-arts in pruning large networks. For instance, CWP improves the accuracy of ResNet56 on CIFAR-10 dataset by 0.32\% aftering removing 64.11\% FLOPs, and prunes 87.75\% FLOPs of ResNet50 on ImageNet dataset with only 0.93\% Top-1 accuracy loss.
\end{abstract}




\maketitle

\section{Introduction}
Convolutional neural networks have shown excellent performance in computer vision tasks~\cite{dong2015image,girshick2014rich,xu2020accelerated,zhao2021adaptive, bosquet2023full, wu2022cross, ma2022regionwise} such as object detection and image recognition. To make the network perform well in various tasks, the size of network has been continuously increased, which results in extensive computation consumption. Deploying such large networks to edge devices is impossible due to limited computing resources. To cope with this issue, model compression has been introduced to obtain compact and efficient subnetworks with little damage to the model performance.

\begin{figure*}[ht]
	\centering
	\includegraphics[width=0.9\textwidth]{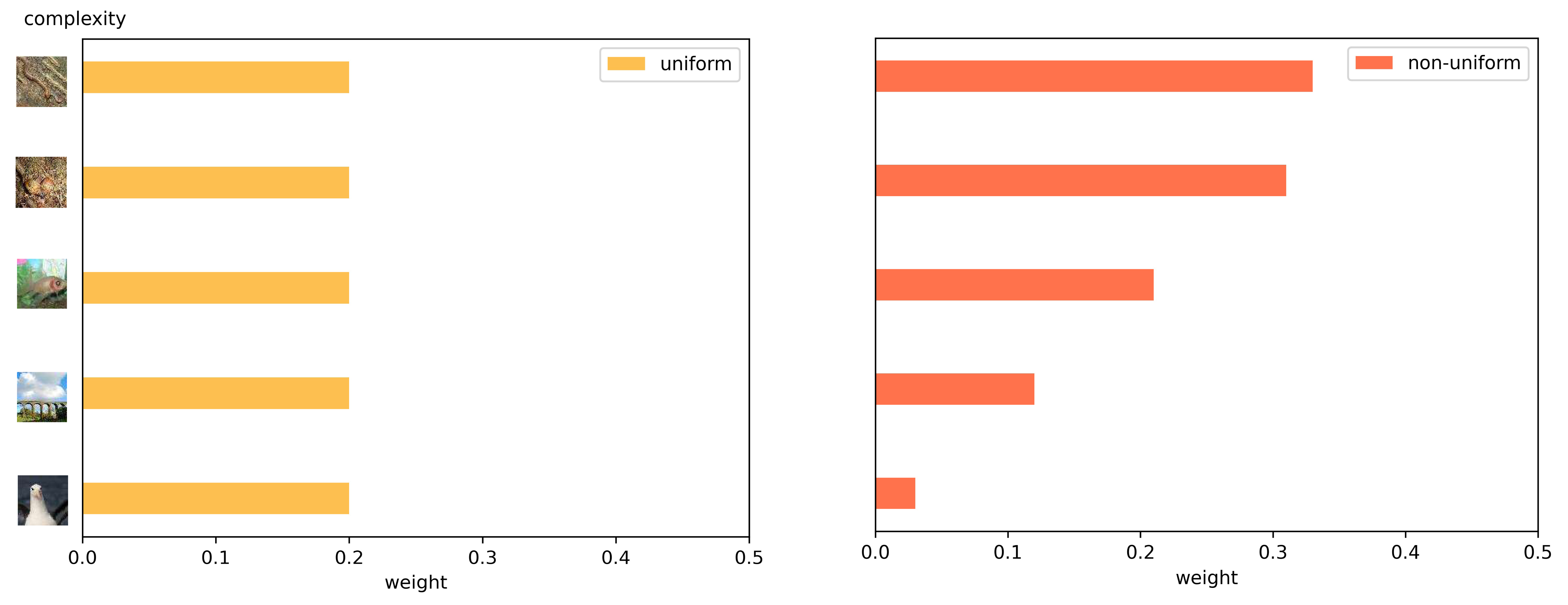}
	\caption{Uniform weights used in existing methods, and non-uniform weights proposed in CWP to measure the filter importance. CWP gives higher weights to complex instances that are difficult to classify.}
	\label{fig1}
\end{figure*}

There are many excellent model compression methods including knowledge distillation~\cite{xu2019lightweightnet,hinton2015distilling}, low-rank decomposition~\cite{yu2017compressing}, and network quantization~\cite{wan2020deep, wei2022qdrop}. They all have their own advantages and disadvantages while complement each other. A plenty of recent approaches have focused on another model compression method - \textbf{Pruning}. Pruning can be divided into structured pruning and unstructured pruning. Unstructured pruning operates on weights or connections of network, and requires specific hardware devices or libraries to speed up computation after pruning. Structured network pruning is a widely used model compression technique, it aims to find and remove redundant structures from a pre-trained or baseline network. Current differentiable pruning methods~\cite{lin2018accelerating,you2019gate,NEURIPS2019_b51a15f3,gao2020discrete,kim2020plug,kang2020operation,tang2021manifold} often introduce channel-wise masks (or gates) to indicate pruning status of each filter, and employ sparsity constraints such as $\ell_1$ regularizer to make some gates or masks approach zero. In these methods, a threshold on the masks is explored to separate the filters into two parts for removing the less important one, and fine-tuning of the pruned network is required to compensate the accuracy loss. However, conventional $\ell_1$ or $\ell_2$ norms do not promise a complete separation of important filters from redundant ones, and the threshold is usually selected based on target FLOPs reduction rate, which may heavily affect the accuracy of the pruned network due to discretization gap between training and evaluation. To cope with this issue, several recently published works~\cite{zhuang2020neuron, Guo_2021_ICCV, pmid35571691} have proposed to polarize the masks during network training. For instance, Zhuang \etal~\cite{zhuang2020neuron} employs a $\ell_1$ norm based polarization regularizer to divide the filters into two parts, and Guo \etal~\cite{Guo_2021_ICCV} also uses gates with differentiable polarization to prune channels. These newly introduced regularizations help distinguish between redundant and important filters since there are often suitable spots that completely separate the two sets of filters. Given the advantage of polarization, there still needs further works to maximize the margin between the masks of important and redundant filters, thus better preserve the model capability. 

Recent dynamic pruning methods consider instance complexity when generating instance-wise subnetworks~\cite{tang2021manifold}. The reported results show hard inputs generally require more complex subnetworks to make an accurate prediction. Unlike dynamic pruning that generates instance-wise sparsity at run-time~\cite{tang2021manifold,Chen_2019_CVPR,Lin2020Dynamic,Li_2021_CVPR}, static pruning obtains a fixed subnetwork applied to all inputs, and instance complexity may play an important role when determining which filters should be removed. However, existing static pruning methods often implicitly assume a uniform contribution of different instances when measuring filter importance, and not yet fully investigate the effects of instance complexity on channel pruning. How to incorporate instance complexity into the assessment of filter importance is still an open problem in static network pruning.

\begin{figure*}[ht]
	\centering
	\includegraphics[width=0.95\textwidth]{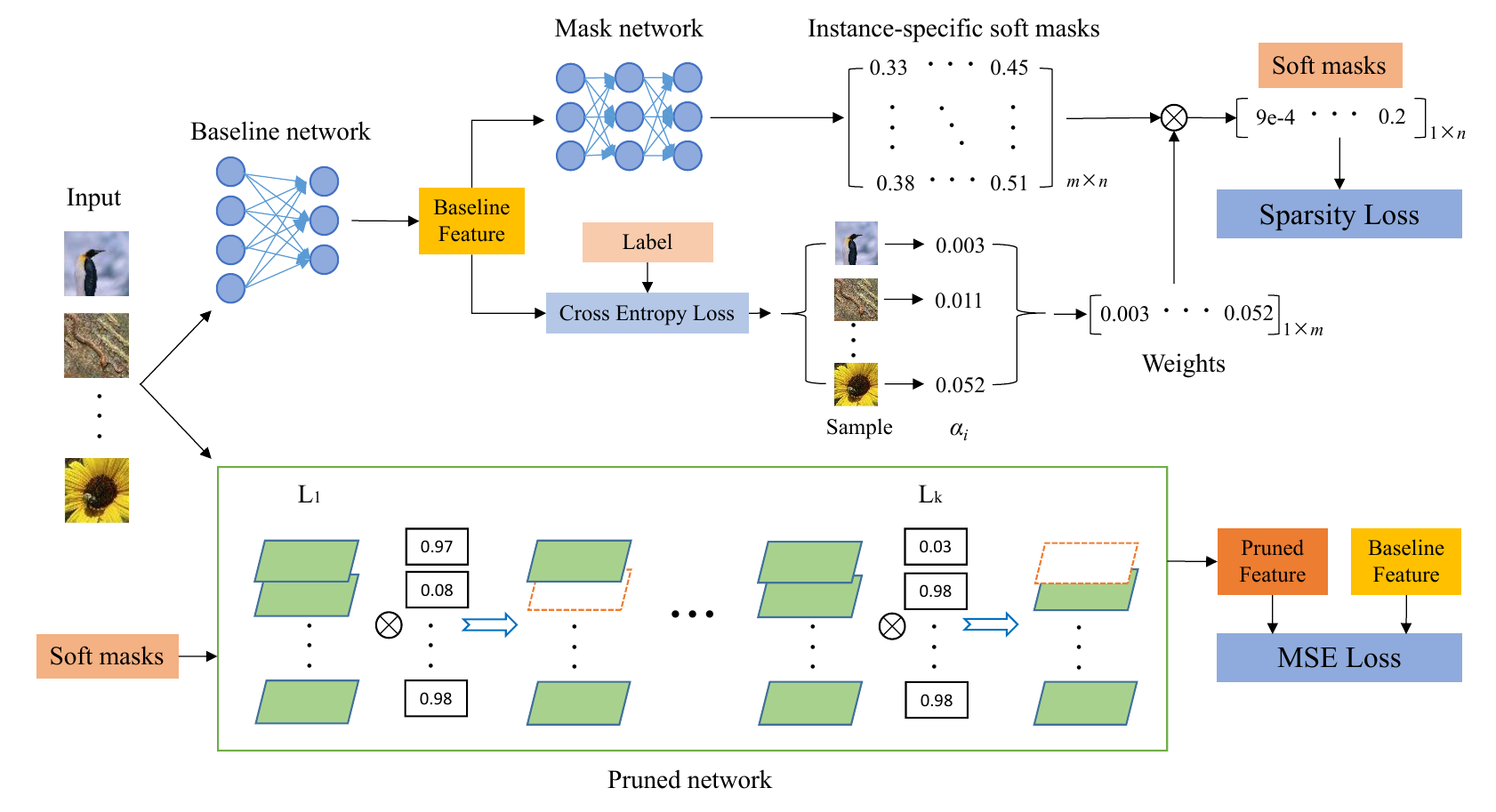}
	\caption{An illustration of CWP framework. CWP measures the weighted sum of instance-specific soft masks to represent the filter importance, and use the weighted soft masks to scale the feature maps of the pruned network. The instance-specific soft masks are learned by passing baseline features of each instance through a fully connected network (Mask network), and the weight of each instance is calculated with the cross entropy loss. The model parameters are jointly optimized by minimizing MSE loss between baseline features and pruned features, and sparsity loss related to the weighted soft masks.}
	\label{fig2}
\end{figure*}

In this paper, we propose instance \underline{C}omplexity \underline{W}eighted soft masks for static network \underline{P}runing (CWP) to prune a large pre-trained (baseline) network. Similar to related approaches, CWP employs a soft mask with value between 0 and 1 to represent the importance of each filter, and removes the filters with masks close to zero. Our method also shows two new features when compared to the existing methods: 1) to model the non-uniform contribution of different inputs to filter importance, we define the importance scores of all filters with a weighted sum of instance-specific soft masks, where the weight of each instance is closely related to the instance complexity and higher weights are given to hard instances (as shown in Figure~\ref{fig1}); 2) to clearly separate the important filters from redundant ones, we introduce a regularizer on the soft masks of filters to encourage polarization of the masks, such that a sweet spot can be easily found to divide the filters into two parts with a large margin between them. These new modeling features enable CWP to yield better pruning results than the state-of-the-art (SOTA) methods. An illustration of CWP framework is given in Figure~\ref{fig2}.

Our main contributions are threefold:
\begin{itemize}
	\item By exploiting the difference of instance complexity, non-uniform contribution of inputs to filter importance is introduced to give higher weights for hard instances, which reduces the performance gap between the pruned and original networks.
	\item By employing a new regularizer on the soft masks of filters, the masks of important filters are pushed towards 1 and those of redundant filters are pushed towards 0, which effectively relieves the discretization gap between training and evaluation.
	\item Extensive experiments on multiple datasets and popular network architectures demonstrate CWP outperforms the SOTAs by pruning more FLOPs with less accuracy loss.
\end{itemize}

\section{Related Works}

\subsection{Network pruning}
Network pruning methods can be divided into unstructured and structured pruning. Unstructured pruning~\cite{dong2017learning,han2015learning,chijiwa2021pruning} often exploits sparsity-oriented penalty on network weights to remove unimportant elements in weight tensors, and is not hardware friendly since it requires specially designed library for accelerating multiplication between sparse matrices. By comparison, structured pruning~\cite{he2018amc,9102834,huang2021rethinking,chen2021only} removes regular structures such as channels, filters, neurons and layers of network under specific pruning criteria. Early works mainly adopt an iterative process to generate and evaluate subnetworks. For instance, AMC~\cite{he2018amc} employs reinforcement learning to reason reserving rate of each layer under a layer-by-layer manner. To improve subnetwork search efficiency, Cao \etal~\cite{cao2018learnable} uses Bayesian optimization algorithm to iteratively sample subnetworks for evaluation. There are also methods that calculate importance scores of the filters to guide channel pruning~\cite{liebenwein2019provable,chin2020towards,lin2020hrank}. For instance, HRank~\cite{lin2020hrank} defines the importance score of a filter as the rank of feature map matrix, and CHIP~\cite{sui2021chip} proposes to use channel independence to measure the filter importance. In recent years, differentiable network pruning~\cite{gao2020discrete,kim2020plug,kang2020operation,tang2021manifold} has been widely explored to prune network under an end-to-end manner. These methods often attach learnable scaling factors or masks behind channels, and introduce sparsity-oriented regularizers on the scaling factors to identify redundant channels. For instance, GAL~\cite{lin2019towards} uses adversarial learning to reason the soft masks, DMCP~\cite{guo2020dmcp} defines retaining probability of each channel under a Markov process, DPFPS~\cite{ruan2021dpfps} conducts dynamic and progressive filter pruning by employing a sparsity regularizer, and FTWT~\cite{elkerdawy2022fire} uses a self-supervised approach to predict binary gating for dynamic pruning. Our method is also based on learning of the soft masks, and differs with existing methods on two aspects: non-uniform contribution of input instances to filter importance is considered in CWP; a new regularizer on the soft masks is used to reduce the accuracy loss after pruning.

\subsection{Neural architecture search}
Neural Architecture Search (NAS) has attracted more and more attention for its capability of automatically finding a high-performance architecture from a large search space. Early NAS works are based on either reinforcement learning~\cite{zhong2018practical,pham2018efficient} or revolutionary algorithm~\cite{real2017large,liu2017hierarchical}. To continuously deal with the search space, differentiable architecture search~\cite{liu2018darts} is proposed to facilitate optimization techniques such as gradient descent to be employed for finding the optimal architecture. Our method can also be treated as a NAS process where the channel-wise masks form the search space. Compared with existing NAS methods, our work introduces the non-uniform contribution of instances to filter importance.

\section{Proposed CWP Framework}

\subsection{Notations and Preliminaries}

Given a training dataset $\{\mathbf{x}_i,\mathbf{y}_i\}_{i=1}^N$, we aim to get a well performed pruned network $\mathcal{P}$ from the baseline network $\mathcal{B}$. Here $\mathbf{x}_i$ denotes input image and $\mathbf{y}_i$ is a one-hot encoding vector to represent the label of $\mathbf{x}_i$. The pruned network $\mathcal{P}$ has the same structure to $\mathcal{B}$. For channel pruning, the output of each filter in $\mathcal{P}$ is multiplied by a soft mask before fed into next layer, and all masks are jointly optimized to find and remove filters with mask close to zero. Suppose there are a total of $n$ filters in $\mathcal{P}$, the feature map from the $k$-th filter is represented by $\mathbf{F}_k \in \mathbf{R}^{W_k \times H_k}$ ($W_k$ is the height and $H_k$ is the width of the feature map), and the channel-wise masks are denoted by a $n$-dimensional vector $\mathbf{m}$ of which the $k$-th element is in [0, 1] and used to scale $\mathbf{F}_k$. Here $\mathbf{m}$ is treated as the importance scores of the filters. For each input image $\mathbf{x}_i$, the feature map outputted by the last FC layer is denoted by $\hat{\mathbf{y}}_i=f(\mathbf{x}_i;W_b)$ for the baseline network and $\hat{\mathbf{g}}_i=f(\mathbf{x}_i,\mathbf{m};W_p)$ for the pruned network. Here $W_b$ and $W_p$ refer to the weights of the baseline and pruned networks, respectively. Network pruning can be formulated as the following optimization problem:
\begin{equation}
	\label{formula:f1}
	\min_{\mathbf{m},W_p} \mathcal{L}=\sum_{i=1}^{N} \mathcal{L}_{mse}(\hat{\mathbf{y}}_i,\hat{\mathbf{g}}_i)+\lambda_1 R(W_p)+\lambda_2 R(\mathbf{m})
\end{equation}
where $\mathcal{L}_{mse}$ denotes mean squared error (MSE), $R(W_p)$ is a regularizer (e.g. $\ell_2$-norm) applied to the weights of the pruned network, $R(\mathbf{m})$ represents a sparsity regularizer on $\mathbf{m}$, coefficients $\lambda_1$ and $\lambda_2$ control the weights of $R(W_p)$ and $R(\mathbf{m})$, respectively. A larger $\lambda_2$ induces sparser masks $\mathbf{m}$ and then yields a more compact pruned network.

\subsection{Definition of filter importance}
Motivated by the concept of instance-aware pruning employed in dynamic network pruning~\cite{NIPS2017_a51fb975,tang2021manifold}, we assume each input sample $\mathbf{x}_i$ characterizes a distribution of filter importance. We use $\mathbf{m}_i=g(\mathbf{x}_i)$ to denote the instance-specific soft masks defined by $\mathbf{x}_i$, here $\mathbf{m}_i$ is a $n$-dimensional vector. Specifically, we model $\mathbf{m}_i$ with a fully connected network $\mathcal{M}$ by following a previous pruning approach~\cite{luo2020autopruner} that projects layer-wise tensor into a $C$-dimensional binary masks through a fully connected layer (here $C$ is the number of channels). Instead of directly using $\mathbf{x}_i$ as input, we fed the baseline feature $\hat{\mathbf{y}}_i$ into $\mathcal{M}$ to output the masks $\mathbf{m}_i$, i.e. $\mathbf{m}_i=g(\hat{\mathbf{y}}_i;W_m)$, here $W_m$ denotes the weights of the mask network. This is reasonable as $\hat{\mathbf{y}}_i$ is the high-level representation of $\mathbf{x}_i$ and encompasses more useful features for target tasks. The importance scores of the filters are then defined as the weighted sum of instance-specific soft masks:
\begin{equation}
	\label{formula:f2}
	\mathbf{m}=\sum_{i=1}^{N} \alpha_i g(\hat{\mathbf{y}}_i;W_m)=\sum_{i=1}^{N} \alpha_i g(f(\mathbf{x}_i;W_b);W_m)
\end{equation}
where $\alpha_i$ denotes the weight of the $i$-th input, and is set to $\frac{1}{N}$ when assuming a uniformly distributed contributions of all instances. Here we assume that complex instances require stronger subnetworks than simple instances to extract valid feature information, and should be given more attention when measuring filter importance. To consider non-uniform complexity of the inputs, we propose to give higher weights to hard instances that are more difficult to classify. Following the approach adopted in~\cite{tang2021manifold}, we use cross entropy (CE) loss $\mathcal{L}_{ce}$ to define the complexity of each instance. The $\alpha_i$ is defined as follows:

\begin{equation}
	\label{formula:f3}
	\alpha_i=\frac{\mathcal{L}_{ce}(\mathbf{y}_i,\hat{\mathbf{y}}_i)}{\sum_{j}\mathcal{L}_{ce}(\mathbf{y}_j,\hat{\mathbf{y}}_j)}
\end{equation}

Given above definitions, the inferred $\mathbf{m}$ in Eq.~\ref{formula:f2} is a weighted average of instance-specific soft masks, and used to scale the output features of filters for all instances. The gradient of the loss function with respect to $W_m$ can be decomposed into:
\begin{equation}
	\label{formula:f4}
	\frac{\partial \mathcal{L}}{\partial W_m} = \frac{\partial \mathcal{L}}{\partial \mathbf{m}} \sum_{i=1}^{N} \alpha_i \frac{\partial g(\hat{\mathbf{y}}_i;W_m)}{\partial W_m}
\end{equation}
As complex instances are given higher weights, the weights of mask network $\mathcal{M}$ are updated predominantly with the gradients gathered from complex instances, which encourages complex instances to dominate the learning of the soft masks.

\subsection{Sparsity regularizer on the soft masks}
To determine which filters should be pruned, we need to push the soft masks of redundant filters towards 0 and those of important filters towards 1 to better preserve the model capability after pruning. $\ell_1$-norm has been widely used as a sparsity regularizer~\cite{Liu_2017_ICCV,tang2021manifold}, while it may not promise finding of sweet spots that clearly separate filters into two parts. To cope with this problem, we introduce a new regularizer on the soft masks of filters by following a similar idea in~\cite{zhuang2020neuron}:
\begin{equation}
	\label{formula:f5}
	R(\mathbf{m})=\sum_{k=1}^{n}\textrm{m}_k+t(1-\textrm{var}(\mathbf{m}))
\end{equation}
where $\textrm{var}(\mathbf{m})$ denotes the variance of $\textrm{m}_1$, $\textrm{m}_2$, $\ldots$, $\textrm{m}_n$. The first term is the $\ell_1$-norm which reaches its minimum when all the $\textrm{m}_i$ are zero, and minimizing the second term $1-\textrm{var}(\mathbf{m})$ enables $\textrm{m}_i,1\le i \le n$ to be far away from the mean. Combining these two terms makes it possible to push the soft masks of some filters to 0 and those of remaining filters to 1, such that important filters can be easily distinguished from redundant ones. The parameter $t$ is introduced to balance the two terms.

\subsection{Optimization}
The optimization of the pruned network $\mathcal{P}$ and mask network $\mathcal{M}$ is performed in an end-to-end manner. By combining the MSE loss and regularizers on $W_p$, $\mathbf{m}$ and $W_m$, we obtain the following final objective function that can be optimized under a differentiable manner:
\begin{equation}
	\label{formula:f6}
	\begin{split}
		\min_{W_p,W_m} &\sum_{i=1}^{N} \mathcal{L}_{mse}(\hat{\mathbf{y}}_i,\hat{\mathbf{g}}_i)+\lambda_1 \Vert W_p \Vert_2+\lambda_2 \Vert W_m \Vert_2\\
		&+\lambda_3 \Vert \mathbf{m} \Vert_1+\lambda_4(1-\textrm{var}(\mathbf{m}))
	\end{split}
\end{equation}
where $\lambda_1$ and $\lambda_2$ are set to 5e-4, hyper-parameters $\lambda_3$ and $\lambda_4$ are selected according to the desired FLOPs reduction rate in different experiments. The function can be optimized using conventional mini-batch gradient descent algorithm, and the instance weights defined in Eq.~\ref{formula:f3} are calculated per-batch.

\subsection{Pruning Strategy}
After model training is completed, we obtain the soft masks $\mathbf{m}$ for each batch of inputs, and calculate the average of $\mathbf{m}$ among all batches as the final soft masks. We prune the filters whose soft masks are close to 0, and still need to find a threshold to identify these redundant filters. In ~\cite{zhuang2020neuron}, the threshold is selected by scanning the bimodal distributed scaling factors. Similar to this approach, in our method the filters are clearly separated into two parts one of which locates near to 0 and another locates near to 1 due to the polarization effect, therefore we do not need to explore a threshold on the mask values and simply use 0.5 as a split point to divide the filters into two groups. Bases on this pruning strategy, the filters with soft masks near to 0 can be automatically removed. After pruning, the network is fine-tuned on training data to compensate the accuracy loss.

\section{Experiments}
\subsection{Experiment Setting}
We evaluate our method on three popular datasets including CIFAR-10~\cite{krizhevsky2009learning}, CIFAR-100~\cite{krizhevsky2009learning} and ImageNet~\cite{russakovsky2015imagenet}. CIFAR-10 contains 60k images with size of $32\times32$ for 10 classes, we use 50k images as training set and remaining 10k images as test set. CIFAR-100 has 100 classes with 500 training images and 100 test images per class. The large-scale dataset ImageNet contains 1k classes, including 1.28 million images as training set and 50k images as test set. On CIFAR-10 dataset, we evaluatet the proposed method on ResNet32, ResNet56 and ResNet110 models~\cite{he2016deep}. On CIFAR-100 dataset, we evaluate our method on ResNet32 and ResNet56 models. On ImageNet dataset, we assess our proposed method on ResNet50 following the existing methods.

\begin{table*}[tb]
	\centering
	\caption{Compare the results of ResNet32, ResNet56 and ResNet110 on CIFAR-10 dataset. ACC↓ represents the percent of accuracy drop after pruning. Flops↓ represents FLOPs reduction rate. Best results are bolded.}
	\label{tab1}
	\begin{tabular}{c|c|cccc} 
		\hline
		\begin{tabular}[c]{@{}c@{}}\\Network\end{tabular} & Method & \begin{tabular}[c]{@{}c@{}}Baseline \\Acc(\%)\end{tabular} & \begin{tabular}[c]{@{}c@{}}Pruned\\Acc(\%)\end{tabular} & \begin{tabular}[c]{@{}c@{}}Acc↓\\(\%)\end{tabular} & \begin{tabular}[c]{@{}c@{}}FLOPs↓\\(\%)\end{tabular}  \\ 
		\hline
		\multirow{7}{*}{ResNet32} & FPGM ~\cite{2018Filter} & 92.63 & 92.82 & -0.19 & 53.20 \\
		& LFPC ~\cite{he2020learning} & 92.63 & 92.12 & 0.51 & 52.60 \\
		& Wang et al. ~\cite{wang2021accelerate} & 93.18 & 93.27 & -0.09 & 49.00 \\
		& \textbf{CWP(ours)} & 92.94 & 93.73(±0.04) & \textbf{-0.79(±0.04)} & \textbf{56.81} \\ 
		\cdashline{2-6}
		& MainDP ~\cite{tang2021manifold} & 92.66 & 92.15 & 0.51 & 63.20 \\
		& LRF ~\cite{joo2021linearly} & 92.49 & 92.54 & -0.05 & 62.00 \\
		& \textbf{CWP(ours)} & 92.94 & 93.64(±0.02) & \textbf{-0.7(±0.02)} & \textbf{64.40} \\ 
		\hline
		\multirow{12}{*}{ResNet56} & FPGM ~\cite{2018Filter} & 93.59 & 93.26 & 0.33 & 52.60 \\
		& HRank ~\cite{lin2020hrank} & 93.26 & 93.17 & 0.09 & 50.00 \\
		& LFPC ~\cite{he2020learning} & 93.59 & 93.34 & 0.25 & 52.90 \\
		& DMC ~\cite{gao2020discrete} & 93.62 & 93.69 & -0.07 & 50.00 \\
		& SRR-GR ~\cite{wang2021convolutional} & 93.38 & 93.75 & -0.37 & 53.80 \\
		& SCP ~\cite{kang2020operation} & 93.69 & 93.23 & 0.46 & 51.50 \\
		& DPFPS ~\cite{ruan2021dpfps} & 93.81 & 93.20 & 0.61 & 52.86 \\
		& Wang et al. ~\cite{wang2021accelerate} & 93.69 & 93.76 & -0.07 & 50.00 \\
		& $FTWT_{J}$ ~\cite{elkerdawy2022fire} & 93.66 & 92.28 & 1.38 & 54.00 \\
		& \textbf{CWP(ours)} & 93.25 & 93.68(±0.03) & \textbf{-0.43(±0.03)} & \textbf{58.36} \\ 
		\cdashline{2-6}
		& LRF ~\cite{joo2021linearly} & 93.45 & 93.73 & -0.28 & 62.40 \\
		& MainDP ~\cite{tang2021manifold} & 93.70 & 93.64 & 0.06 & 62.40\\
		& $FTWT_{D}$ ~\cite{elkerdawy2022fire} & 93.66 & 92.63 & 1.03 & 66.00 \\
		& DNAL ~\cite{guo2022differentiable} & 94.15 & 93.20 & 0.95 & \textbf{70.00} \\
		& \textbf{CWP(ours)}& 93.25 & 93.57(±0.03) & \textbf{-0.32(±0.03)} & 64.11 \\ 
		\hline
		\multirow{7}{*}{ResNet110} & FPGM ~\cite{2018Filter} & 93.68 & 93.85 & -0.17 & 52.30 \\
		& TAS ~\cite{dong2019network} & 94.97 & 94.33 & 0.64 & 53.00 \\
		& \textbf{CWP(ours)} & 93.60 & 93.93(±0.02) & \textbf{-0.33(±0.02)} & \textbf{55.38} \\ 
		\cdashline{2-6}
		& LRF ~\cite{joo2021linearly} & 93.76 & 94.34 & \textbf{-0.58} & 62.60 \\
		& HRank ~\cite{lin2020hrank} & 93.50 & 92.65 & 0.85 & 68.60 \\
		& LFPC ~\cite{he2020learning} & 93.68 & 93.79 & -0.11 & 60.30 \\
		& \textbf{CWP(ours)} & 93.60 & 93.86(±0.04) & -0.26(±0.04) & \textbf{68.73} \\
		\hline
	\end{tabular}
\end{table*}

\textbf{Implementation Details:} The mask network $\mathcal{M}$ consists of three FC layers and a sigmoid layer, and dimension of the output is equal to the number of filters of the baseline network. The baseline traning is set to run 200 epochs on CIFAR-10, CIFAR-100 and ImageNet. The initial learning rate is set to 0.1, and batch size is set to 128. The learning rate is multiplied by 0.1 after every 30 epochs. We use the mini-batch gradient descent method for optimization with momentum of 0.9 and weight decay of 0.0004. In the pruning stage, for different baselines we train on different datasets for 30 epochs with a learning rate of 0.01 and multiply the learning rate by 0.1 after every 5 epochs. The pruned networks are fine-tuned with the same parameters as used in training the baselines, except that the initial learning rate is set to 0.01. We conduct 5 rounds of experiments for each experimental setup to check robustness of the pruning results. The models are implemented in PyTorch and all experiments are executed on a RTX TITAN GPU.

\textbf{Hyperparameter:} During pruning procedure, we set the two hyper-parameters $\lambda_3$ and $\lambda_4$ to achieve target FLOPs reduction rate. On CIFAR-10 and CIFAR-100 datasets, we fix the value of $\lambda_3$ to 0.001, then explore the value of $\lambda_4$ in range of [1, 20] to reach the target FLOPs reduction rate. On ImageNet dataset, we also set $\lambda_3$ to 0.001 and search $\lambda_4$ in [20, 80] to get target FLOPs reduction rate.

\begin{table*}[tb]
	\centering
	\caption{Compare the results of ResNet32 and ResNet56 on CIFAR-100 dataset. ACC↓ represents the percent of accuracy drop after pruning. Flops↓ represents FLOPs reduction rate.}
	\label{tab2}
	\begin{tabular}{c|c|cccc} 
		\hline
		Network & Method & \begin{tabular}[c]{@{}c@{}}Baseline\\Acc(\%)\end{tabular} & \begin{tabular}[c]{@{}c@{}}Pruned\\Acc(\%)\end{tabular} & \begin{tabular}[c]{@{}c@{}}Acc↓\\(\%)\end{tabular} & \begin{tabular}[c]{@{}c@{}}FLOPs↓\\(\%)\end{tabular}  \\ 
		\hline
		\multirow{5}{*}{ResNet32} & LCCL ~\cite{dong2017more} & 70.08 & 67.39 & 2.69 & 37.5 \\
		& SFP ~\cite{kang2020operation} & 69.77 & 68.37 & 1.40 & 41.5 \\
		& FPGM ~\cite{2018Filter} & 69.77 & 68.52 & 1.25 & 41.5 \\
		& TAS ~\cite{dong2019network} & 70.6 & 66.94 & 3.66 & 41 \\
		& \textbf{CWP(ours)} & 69.51 & 70.45(±0.04) & \textbf{-0.94(±0.04)} & \textbf{51.47} \\ 
		\hline
		\multirow{5}{*}{ResNet56} & LCCL ~\cite{dong2017more} & 71.33 & 67.39 & 2.96 & 39.3 \\
		& SFP ~\cite{kang2020operation} & 71.4 & 68.79 & 2.61 & 52.6 \\
		& FPGM ~\cite{2018Filter} & 71.41 & 69.66 & 1.75 & 52.6 \\
		& TAS ~\cite{dong2019network} & 73.18 & 72.25 & 0.93 & 51.3 \\
		& \textbf{CWP(ours)} & 71.34 & 72.13(±0.03) & \textbf{-0.69(±0.03)} & \textbf{55.51} \\
		\hline
	\end{tabular}
\end{table*}

\begin{table*}[tb]
	\centering
	\caption{Compare the results of ResNet50 on ImageNet dataset. Top-1 and Top-5 accuracy denote the different accuracies of baseline. Top-1 pruned and Top-5 pruned accuracy represent the accuracy after pruning. Top-1↓ and Top-5↓ represent the percent of accuracy drop after pruning. Flops↓ represents FLOPs reduction rate.}
	\label{tab3}
	\resizebox{\textwidth}{!}{
		\begin{tabular}{c|c|ccccccc} 
			\hline
			\begin{tabular}[c]{@{}c@{}}\\Network\end{tabular} & Method & \begin{tabular}[c]{@{}c@{}}Top-1\\Acc(\%)\end{tabular} & \begin{tabular}[c]{@{}c@{}}Top-1 Pruned\\Acc(\%)\end{tabular} & \begin{tabular}[c]{@{}c@{}}Top-1↓\\(\%)\end{tabular} & \begin{tabular}[c]{@{}c@{}}Top-5\\Acc(\%)\end{tabular} & \begin{tabular}[c]{@{}c@{}}Top-5 Pruned\\Acc(\%)\end{tabular} & \begin{tabular}[c]{@{}c@{}}Top-5↓\\(\%)\end{tabular} & \begin{tabular}[c]{@{}c@{}}FLOPs↓\\(\%)\end{tabular} \\ 
			\hline
			\multirow{12}{*}{ResNet50}& OTO ~\cite{chen2021only} & 76.10 & 74.70 & 1.4 & 92.90 & 92.10 & 0.8 & 64.50 \\
			& ~OTO* ~\cite{chen2021only} & 76.10 & 75.10 & 1.00 & 92.90 & 92.50 & 0.4 & 64.50 \\
			& ~CHIP ~\cite{sui2021chip} & 76.15 & 75.26 & 0.89 & 92.87 & 92.53 & 0.34 & 62.80 \\
			& ResRep ~\cite{ding2021resrep} & 76.15 & 75.30 & 0.85 & 92.87 & 92.47 & 0.4 & 62.10 \\
			& HRank ~\cite{lin2020hrank} & 76.15 & 71.98 & 4.17 & 92.87 & 91.01 & 1.86 & 62.10 \\
			& DNAL ~\cite{guo2022differentiable} & 75.19 & 72.86 & 2.33 & 92.56 & 91.34 & 1.22 & 64.79 \\
			& AutoPruner ~\cite{luo2020autopruner} & 76.15 & 73.05 & 3.1 & 92.87 & 91.25 & 1.62 & 66.01 \\
			& CHIP ~\cite{sui2021chip} & 76.15 & 73.30 & 2.85 & 92.87 & 91.48 & 1.39 & 76.70 \\
			& Liu et al. ~\cite{liu2021group} & 76.79 & 73.94 & 2.85 & - & - & -  & 75.00 \\
			& DMCP ~\cite{guo2020dmcp} & 76.60 & 74.40 & 2.2 & - & - & - & 73.17 \\
			& Zhuang et al. ~\cite{zhuang2020neuron}& 76.15 & 74.15 & 2.00 & - & - & - & 70.00 \\
			& HRank ~\cite{lin2020hrank} & 76.15 & 69.10 & 7.05 & 92.87 & 89.58 & 3.29 & 76.04 \\
			& DNAL ~\cite{guo2022differentiable} & 75.19 & 70.17 & 5.02 & 92.56 & 89.78 & 2.78 & 78.00 \\
			& \textbf{CWP(ours)} & 76.13 & 75.35(±0.04) & \textbf{0.78(±0.04)} & 92.86 & 92.53(±0.03) & \textbf{0.33(±0.03)} & 73.44 \\
			& \textbf{CWP(ours)} & 76.13 & 75.20(±0.03) & 0.93(±0.03) & 92.86 & 92.45(±0.04) & 0.41(±0.04) & \textbf{87.75} \\
			\hline
	\end{tabular}}
\end{table*}

\subsection{Results on CIFAR-10 and CIFAR-100}
We first evaluate our proposed method on CIFAR-10 dataset by pruning ResNet32, ResNet56 and ResNet110 networks. As shown in Table~\ref{tab1}, we compare CWP to some recently published pruning methods including LFPC~\cite{he2020learning}, LRF~\cite{joo2021linearly}, MainDP~\cite{tang2021manifold}, HRank~\cite{lin2020hrank}, DMC~\cite{gao2020discrete}, DNAL~\cite{guo2022differentiable} and SCP~\cite{kang2020operation}. On ResNet32, our method improves the accuracy by 0.79\% after pruning 56.81\% FLOPs, and outperforms LFPC and FPGM~\cite{2018Filter} at similar FLOPs reduction rates. When 64.4\% FLOPs are pruned, the accuracy improvement of CWP is still better than that of LRF and MainDP. On ResNet56, we include additional competitive methods for comparison. At different target FLOPs reduction rates, CWP prunes more FLOPs and simultaneously yields higher accuracy boost than other methods. For instance, compared with the best-performing existing method SRR-GR~\cite{wang2021convolutional}, our method has better accuracy improvement (0.43\% vs 0.37\%) and higher pruning rate (58.36\% vs 53.80\%). When 64.11\% FLOPs are pruned, CWP improves the accuracy by 0.32\%, and outperforms LRF (0.28\% accuracy improvement), DNAL (0.95\% accuracy loss) and MainDP (0.06\% accuracy loss) at similar FLOPs drop rates. For ResNet110, our method achieves 0.33\% accuracy improvement with 55.38\% FLOPs reduced, while TAS~\cite{dong2019network} results in 0.64\% accuracy loss when pruning similar proportion of FLOPs. CWP also shows advantage in producing more compact subnetworks, it improves the accuracy by 0.26\% after removing 68.73\% FLOPs, and outperforms HRank by a large margin on model accuracy.

We then evaluate our method on CIFAR-100 dataset, and the results are given in Table~\ref{tab2}. For ResNet32, CWP improves the accuracy by 0.94\% with 51.47\% FLOPs reduction rate, while other methods show accuracy loss even at lower FLOPs pruning rates. For instance, the accuracy loss of SFP~\cite{kang2020operation} is as high as 1.4\% when 41.5\% FLOPs are pruned. When pruning 55.51\% FLOPs on ResNet56, CWP still improves accuracy by 0.69\%, and outperforms other methods including LCCL~\cite{dong2017more}, SFP, FPGM~\cite{2018Filter} and TAS~\cite{dong2019network}.

\subsection{Results on ImageNet}
On ImageNet dataset, we evaluate the performance of CWP on ResNet50 network, and compare it with the current advanced methods (Table~\ref{tab3}). Performance metrics such as Top-1 and Top-5 accuracy are adopted for comprehensive comparison. CWP is able to generate more compact networks while better preserve the model accuracy. For instance, Top-1 accuracy only decreases by 0.78\% and Top-5 accuracy only decreases by 0.33\% with 73.44\% FLOPs reduction. By comparison, OTO~\cite{chen2021only} results in 1.00\% loss on Top-1 and 0.4\% loss on Top-5 accuracy with only 64.50\% FLOPs reduction. CWP also yields a highly compact architecture with 87.75\% FLOPs reduction, while causes only 0.93\% Top-1 accuracy drop. At a cost of similar accuracy loss, CHIP~\cite{sui2021chip} only prunes 62.8\% FLOPs. Similar to our method, Zhuang \etal~\cite{zhuang2020neuron} also employs a polarization regularizer on channel-wise scaling factors to find redundant filters, it prunes 70\% FLOPs but results in 2\% accuracy loss. The better performance of CWP when compared to Zhuang \etal~\cite{zhuang2020neuron} may benefit from our modeling of non-uniform contribution of instances to filter importance in CWP. Our method also outperforms recently published methods proposed in~\cite{ding2021resrep} and~\cite{liu2021group}. Taken together, these results suggest CWP scales well to complex dataset.

\subsection{Ablation Study}
\textbf{The effectiveness of non-uniform weights of instances.} Unlike existing methods that implicitly assume a uniform contribution of instances to filter importance, we employ non-uniform weights for instances when measuring the importance of the filters. The weight of each instance is defined based on the CE loss to reflect the complexity of the instance. To verify the effectivess of proposed non-uniform weights, we make a comparison between the pruning results with uniform and non-uniform weights, and the results on CIFAR-10 dataset are shown in Table~\ref{tab4}. With instance complexity considered, CWP is able to significantly enhance the accuracy of the pruned models. For instance, at $\sim$61\% FLOPs reduction rate, only 0.21\% accuracy improvement on ResNet32 is obtained without modeling instance complexity.

\begin{figure*}[tb]
	\centering
	\includegraphics[width=0.98\textwidth]{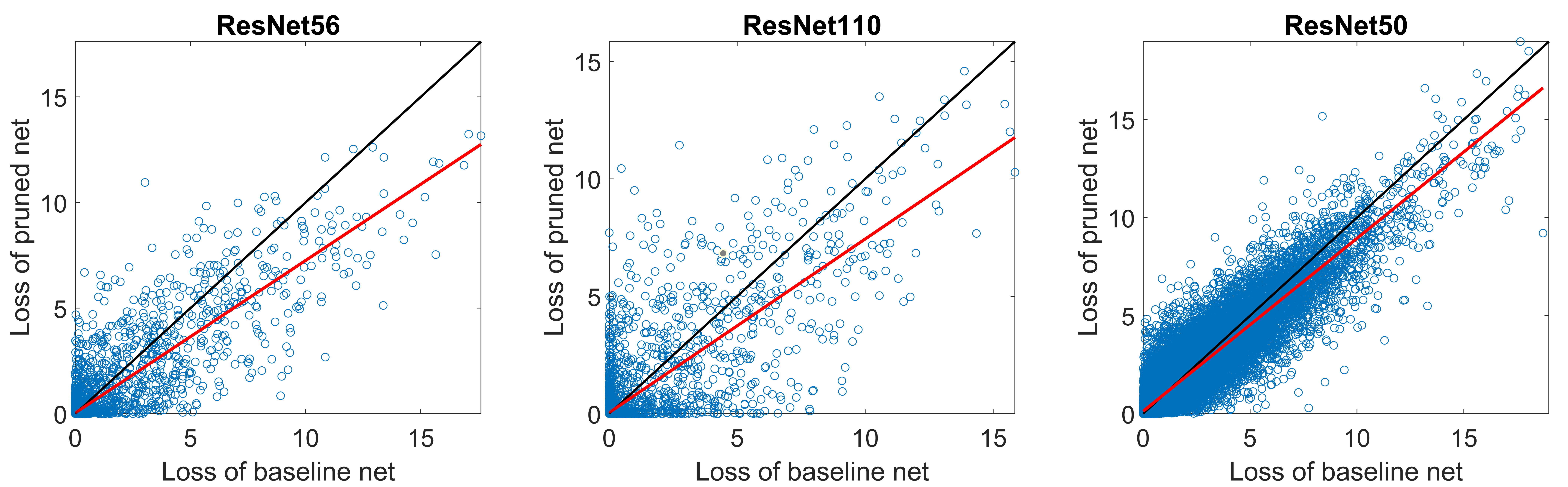}
	\caption{Comparison of CE loss between baseline and pruned networks. For each image in test dataset, CE loss is calculated for the baseline and pruned networks, respectively. Linear regression is employed to analyze the relationship between the loss of pruned and baseline networks (the fitted line is marked by red).}
	\label{fig3}
\end{figure*}

We also compare CE loss of the baseline and pruned networks for ResNet56, ResNet110 and ResNet50 (as shown in Figure~\ref{fig3}). Linear relationship between the loss of pruned network and loss of baseline network is fitted, and the results suggest our proposed method effectively suppresses CE loss for hard instances, thus better preserves the model accuracy. As we give higher weights to hard instances when calculating the soft masks, the network pruning is encouraged to better maintain the prediction capability on complex instances while make correct classification of simple instances.

\begin{figure}[tb]
	\centering
	\subfigure
	{
		\begin{minipage}{3.9cm}
			\centering
			\includegraphics[scale=0.045]{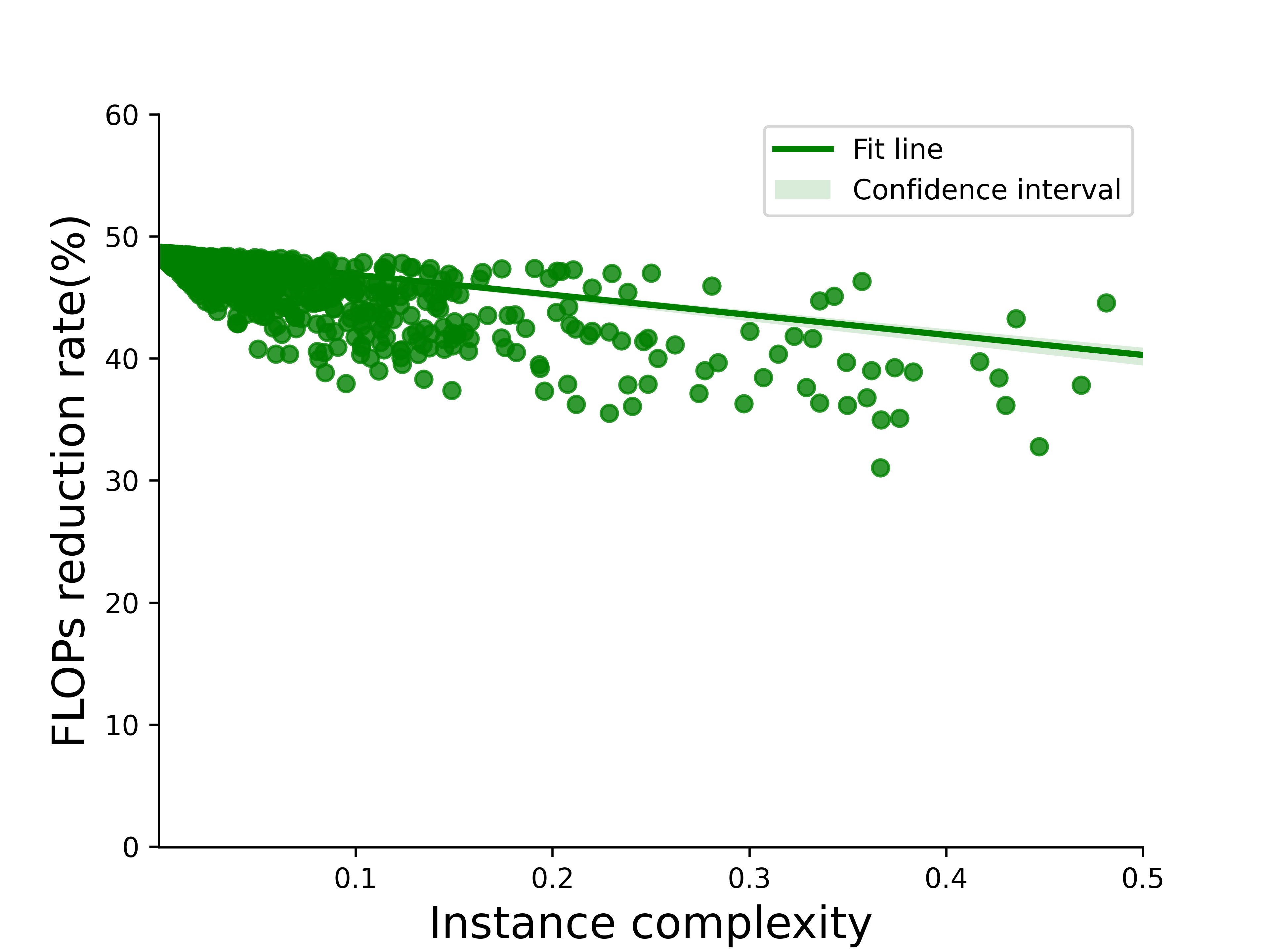}  
		\end{minipage}
	}
	\subfigure
	{
		\begin{minipage}{3.9cm}
			\centering      
			\includegraphics[scale=0.045]{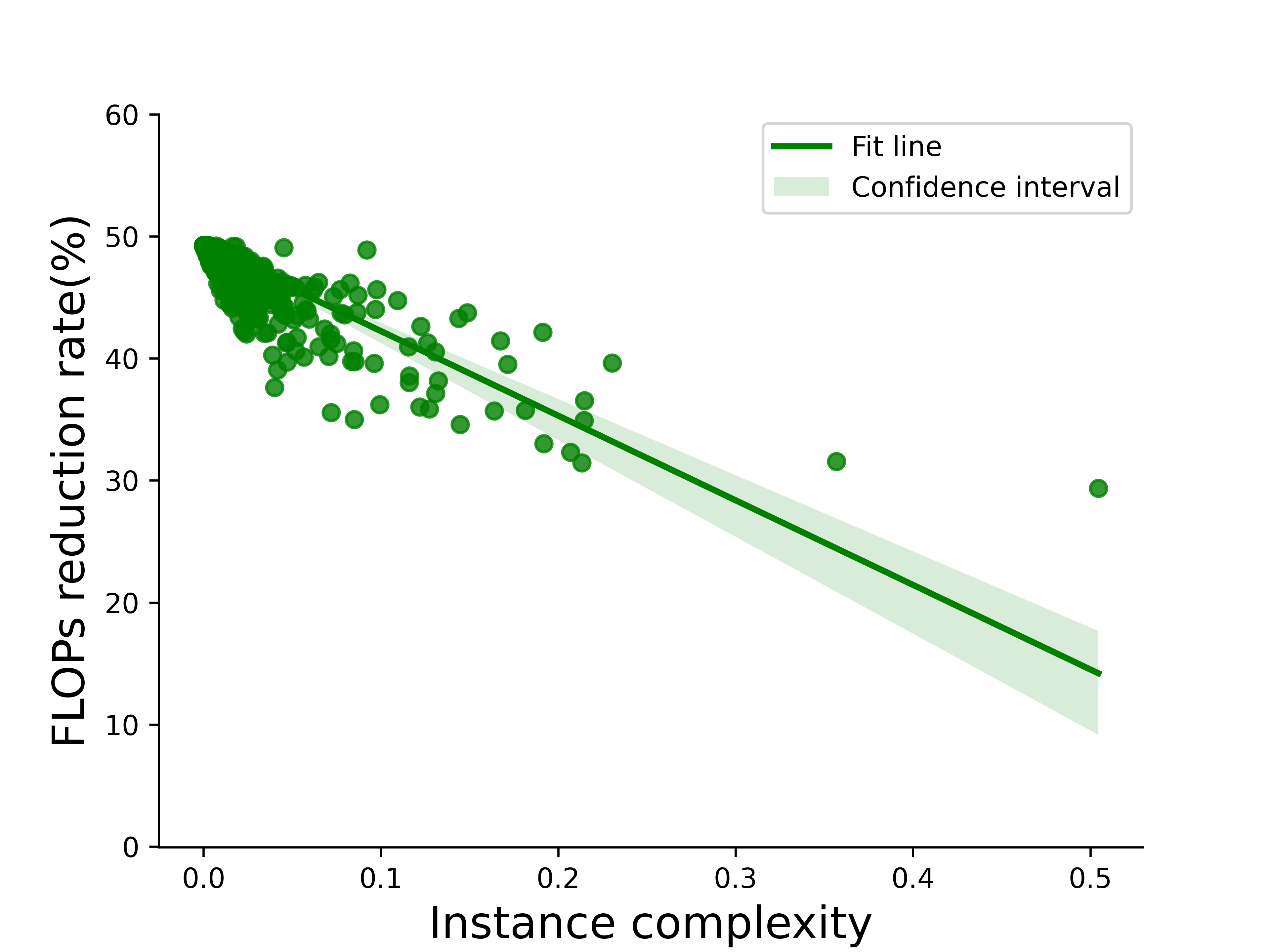}  
		\end{minipage}
	}
	\caption{Analysis of the relationship between instance-specific FLOPs reduction rate and instance complexity after pruning ResNet32 (left subfigure) and ResNet56 (right subfigure) on CIFAR-10 dataset. Based on the instance-specific soft masks outputted by our mask network $\mathcal{M}$, expected FLOPs reduction rate associated with each instance is calculated to verify our assumption that hard instances require more complex subnetworks.}
	\label{fig4}
\end{figure}

To examine if hard instances require more complex subnetworks than simple instances to be accurately classified, we investigate the relationship between per-instance FLOPs reduction rate and instance complexity. The per-instance FLOPs reduction rate is calculated as $\frac{\sum_{j=1}^{n} (1-\mathbf{m}_{ij})F_j}{\sum_{j} F_j}$ (here $n$ is the number of filters, $\mathbf{m}_i$ denotes the instance-specific soft masks defined by the $i$-th instance, and $F_j$ represents the FLOPs of the $j$-th filter). The results in Figure~\ref{fig4} imply hard instances tend to require more complex subnetworks with less FLOPs reduction, and this observation is in concordance with the reported results in dynamic pruning~\cite{tang2021manifold}. The results also suggest our proposed mask network $\mathcal{M}$ effectively captures the complexity difference between hard and simple instances, and tends to assign stronger subnetworks to hard instances.

\begin{table}[tb]
	\centering
	\caption{Performance comparison between uniform and non-uniform weights of instances. The evaluations are conducted on CIFAR-10 dataset to compress ResNet32 and ResNet56 networks. ACC↓ represents the percent of accuracy drop after pruning. Flops↓ represents FLOPs reduction rate.}
	\label{tab4}
	\begin{tabular}{c|c|c|cc} 
		\hline
		Network & Weights & \begin{tabular}[c]{@{}c@{}}Baseline\\Acc(\%)\end{tabular} & \begin{tabular}[c]{@{}c@{}}Acc↓\\(\%)\end{tabular} & \begin{tabular}[c]{@{}c@{}}FLOPs↓\\(\%)\end{tabular}  \\ 
		\hline
		\multirow{2}{*}{ResNet32} & \multirow{1}{*}{non-uniform} & \multirow{2}{*}{92.94} & -0.78 & 62.42 \\ 
		\cdashline{2-2}[1pt/1pt]\cdashline{4-5}[1pt/1pt]
		& \multirow{1}{*}{uniform} & & -0.21 & 60.95 \\ 
		\hline
		\multirow{2}{*}{ResNet56} & \multirow{1}{*}{non-uniform} & \multirow{2}{*}{93.25} & -0.33 & 65.64 \\ 
		\cdashline{2-2}[1pt/1pt]\cdashline{4-5}[1pt/1pt]
		& \multirow{1}{*}{uniform} & & -0.15 & 66.11 \\
		\hline
	\end{tabular}
\end{table}

\begin{table}[tb]
	\centering
	\caption{Comparison of pruning results on ResNet32 and ResNet56 networks with different regularizers on soft masks. ACC↓ represents the percent of accuracy drop after pruning. Flops↓ represents FLOPs reduction rate.}
	\label{tab5}
	\resizebox{\linewidth}{!}{
		\begin{tabular}{c|c|c|cc} 
			\hline
			Network & Regularizer & \begin{tabular}[c]{@{}c@{}}Baseline\\Acc(\%)\end{tabular} & \begin{tabular}[c]{@{}c@{}}Acc↓\\(\%)\end{tabular} & \begin{tabular}[c]{@{}c@{}}FLOPs↓\\(\%)\end{tabular}  \\ 
			\hline
			\multirow{3}{*}{ResNet32} & \multirow{1}{*}{$\ell_1$} & \multirow{3}{*}{92.94} & -0.07 & 62.30 \\ 
			\cdashline{2-2}[1pt/1pt]\cdashline{4-5}[1pt/1pt]
			& \multirow{1}{*}{Zhuang polarization} & & -0.24 & 63.82 \\ 
			\cdashline{2-2}[1pt/1pt]\cdashline{4-5}[1pt/1pt]
			& \multirow{1}{*}{\textbf{Ours}} & & -0.78 & 62.42 \\ 
			\hline
			\multirow{3}{*}{ResNet56} & \multirow{1}{*}{$\ell_1$} & \multirow{3}{*}{93.25} & 0.01 & 60.88 \\ 
			\cdashline{2-2}[1pt/1pt]\cdashline{4-5}[1pt/1pt]
			& \multirow{1}{*}{Zhuang polarization} & & -0.18 & 61.89 \\
			\cdashline{2-2}[1pt/1pt]\cdashline{4-5}[1pt/1pt]
			& \multirow{1}{*}{\textbf{Ours}} & & -0.33 & 65.64 \\
			\hline
	\end{tabular}}
\end{table}

\begin{figure}[tb]
	\centering
	\subfigure 
	{
		\begin{minipage}{3.7cm}
			\centering
			\includegraphics[scale=0.032]{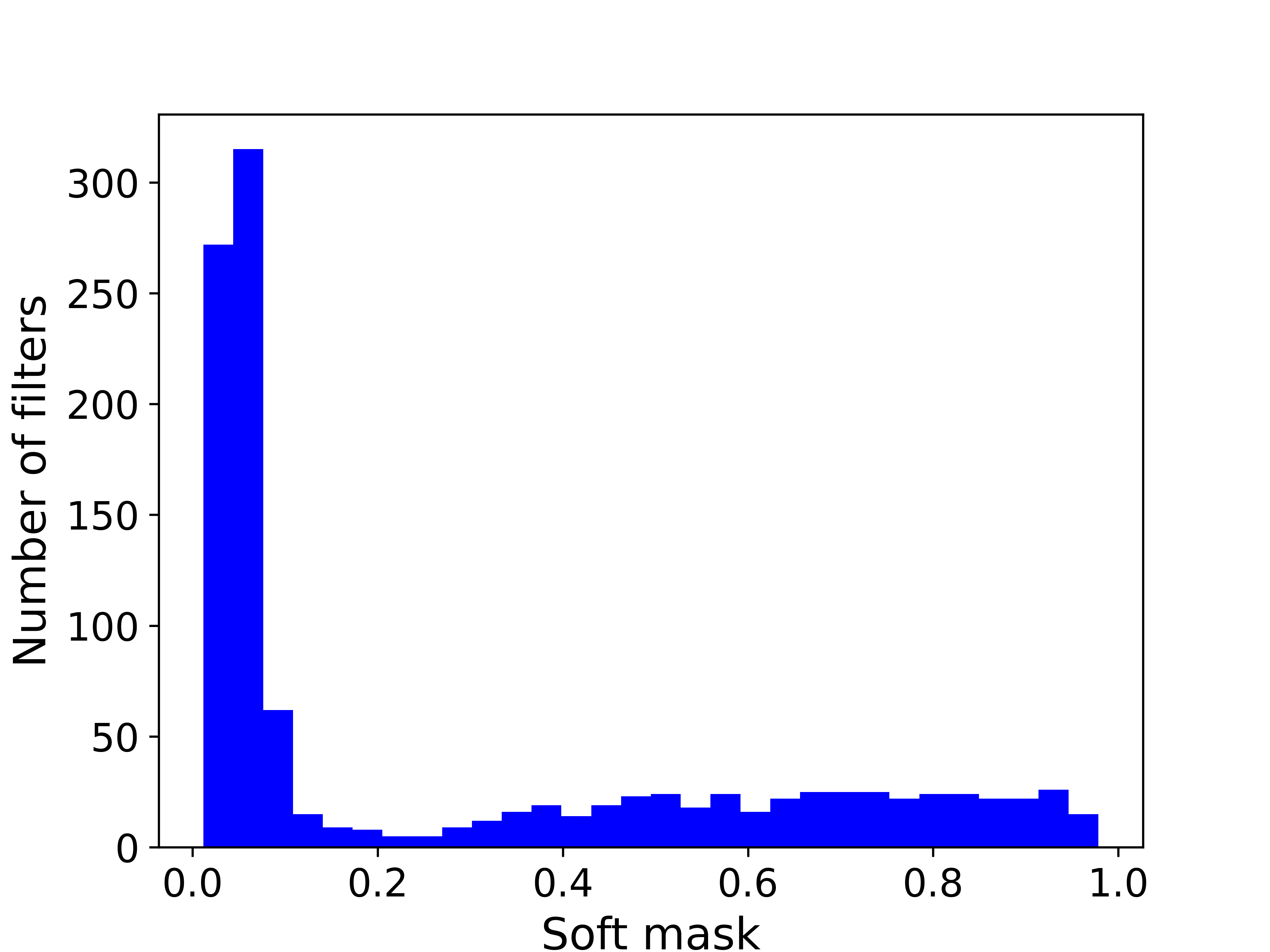}  
		\end{minipage}
	}
	\subfigure
	{
		\begin{minipage}{3.7cm}
			\centering      
			\includegraphics[scale=0.032]{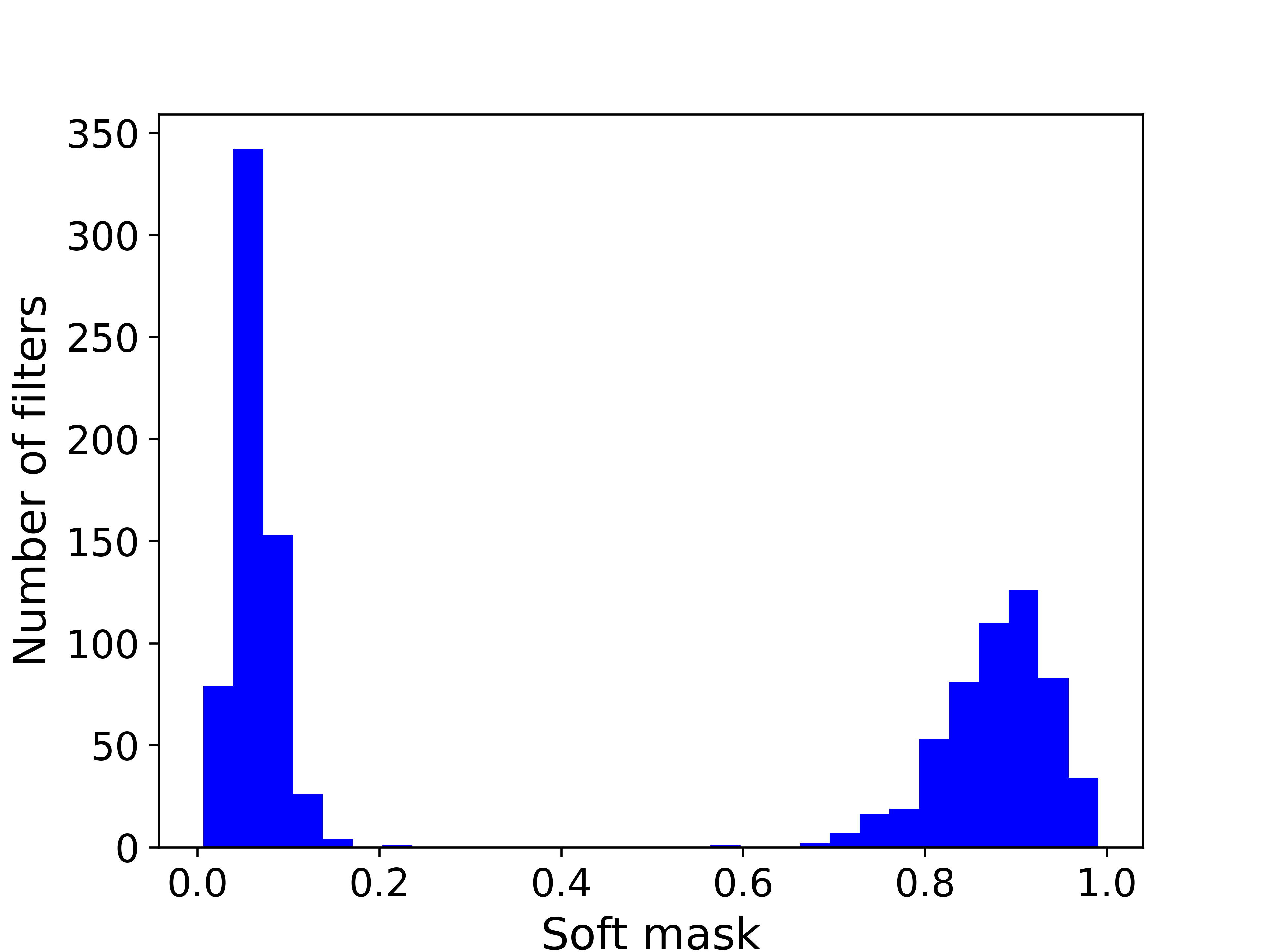}  
		\end{minipage}
	}
	\caption{Histograms of the soft masks are plotted for $\ell_1$ regularizer (left subfigure) and the polarization regularizer proposed in~\cite{zhuang2020neuron} (right subfigure).} 
	\label{fig5}
\end{figure}

\begin{table*}[tb]
	\centering
	\caption{Performance evaluations with either simple or complex instances silenced when measuring the filter importance. ACC↓ represents the percent of accuracy drop after pruning. Flops↓ represents FLOPs reduction rate.}
	\label{tab6}
	\begin{tabular}{c|cccc|c|c|c} 
		\hline
		Network & \begin{tabular}[c]{@{}c@{}}Baseline\\Acc(\%)\end{tabular} & \begin{tabular}[c]{@{}c@{}}Pruned\\Acc(\%) \end{tabular} & \begin{tabular}[c]{@{}c@{}}Acc↓\\(\%)\end{tabular} & \begin{tabular}[c]{@{}c@{}}FLOPs↓\\(\%) \end{tabular} & \begin{tabular}[c]{@{}c@{}}Complex\\instance\end{tabular} & \begin{tabular}[c]{@{}c@{}}Simple\\instance \end{tabular} & Fine tune \\ 
		\hline
		\multirow{4}{*}{ResNet56} & 93.25 & 74.56 & 18.69 & 53.70 & \usym{1F5F8} & \usym{2717} &\multirow{2}{*}{\usym{2717}}\\
		\cdashline{2-7}[1pt/1pt]
		&  93.25 & 66.35 & 26.90 & 52.60 & \usym{2717} & \usym{1F5F8} &  \\
		\cline{2-8}
		&  93.25 & 92.65 & 0.60 & 53.70 & \usym{1F5F8} & \usym{2717} &\multirow{2}{*}{\usym{1F5F8}}\\
		\cdashline{2-7}[1pt/1pt]
		&  93.25 & 91.86 & 1.39 & 52.60 & \usym{2717} & \usym{1F5F8} &  \\
		\hline
	\end{tabular}
\end{table*}

\begin{figure*}[tb]
	\centering
	\subfigure 
	{
		\begin{minipage}{3.7cm}
			\centering
			\includegraphics[scale=0.032]{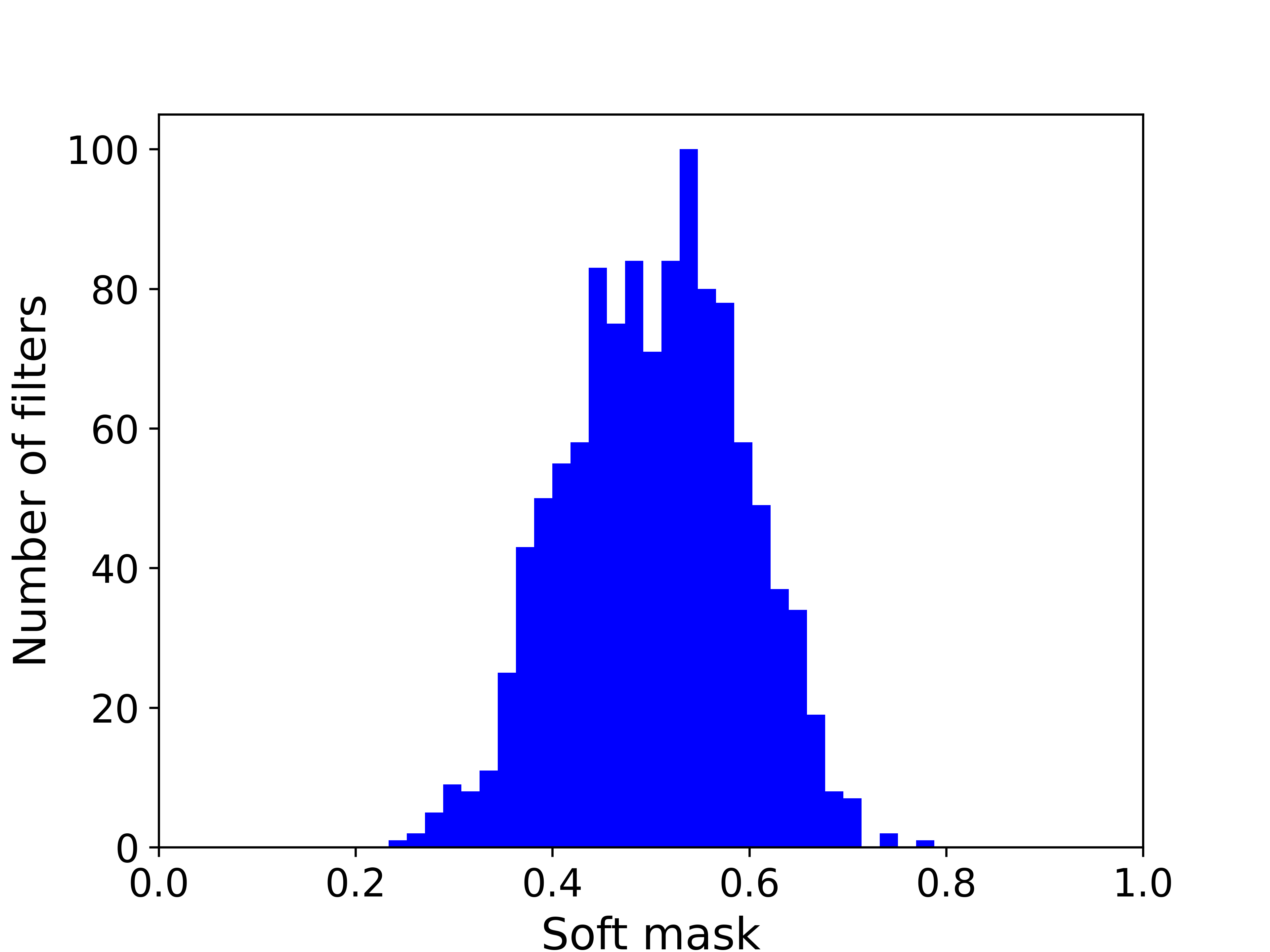}  
		\end{minipage}
	}
	\subfigure
	{
		\begin{minipage}{3.7cm}
			\centering      
			\includegraphics[scale=0.032]{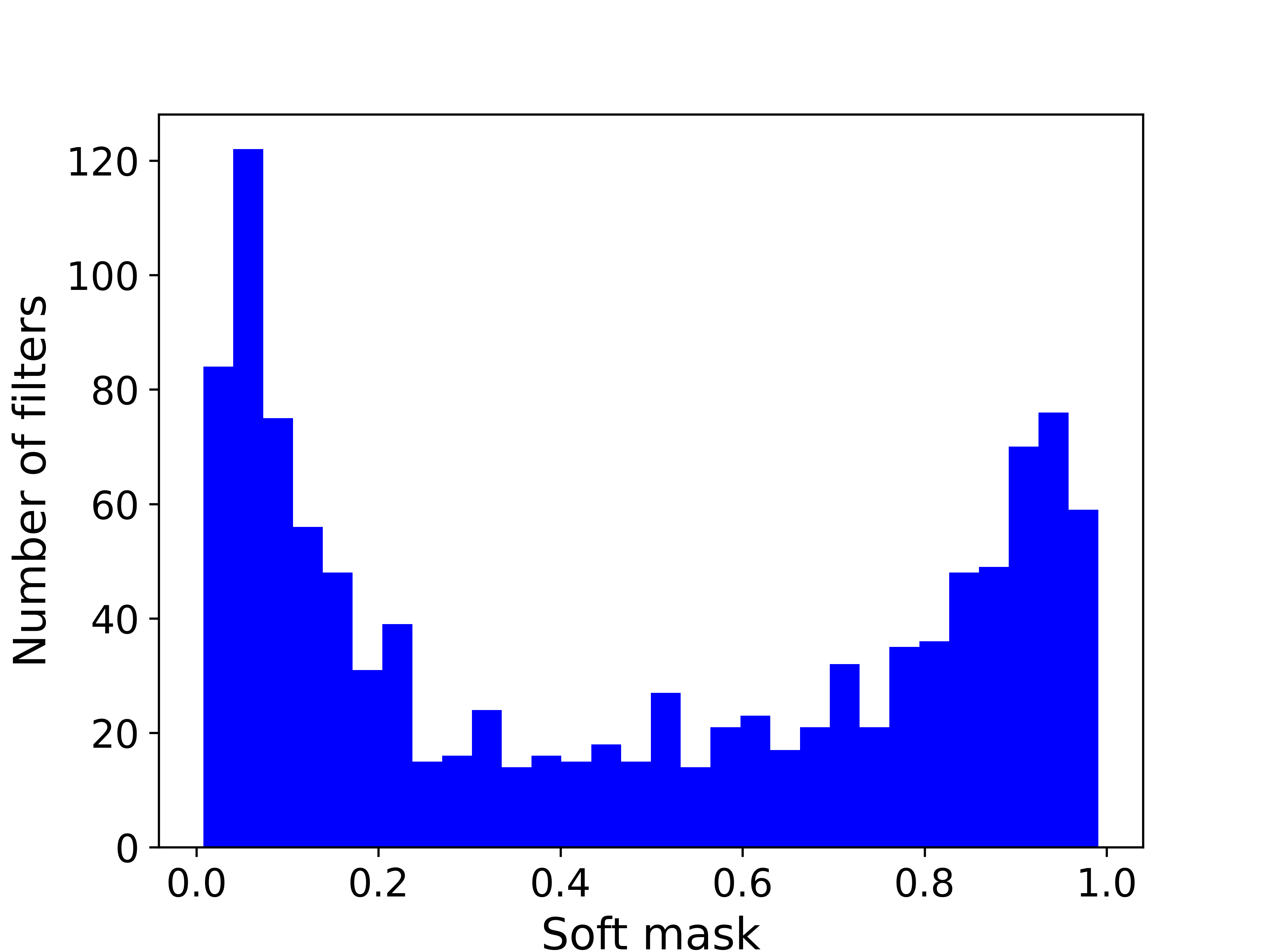}  
		\end{minipage}
	}
	\subfigure
	{
		\begin{minipage}{3.7cm}
			\centering      
			\includegraphics[scale=0.032]{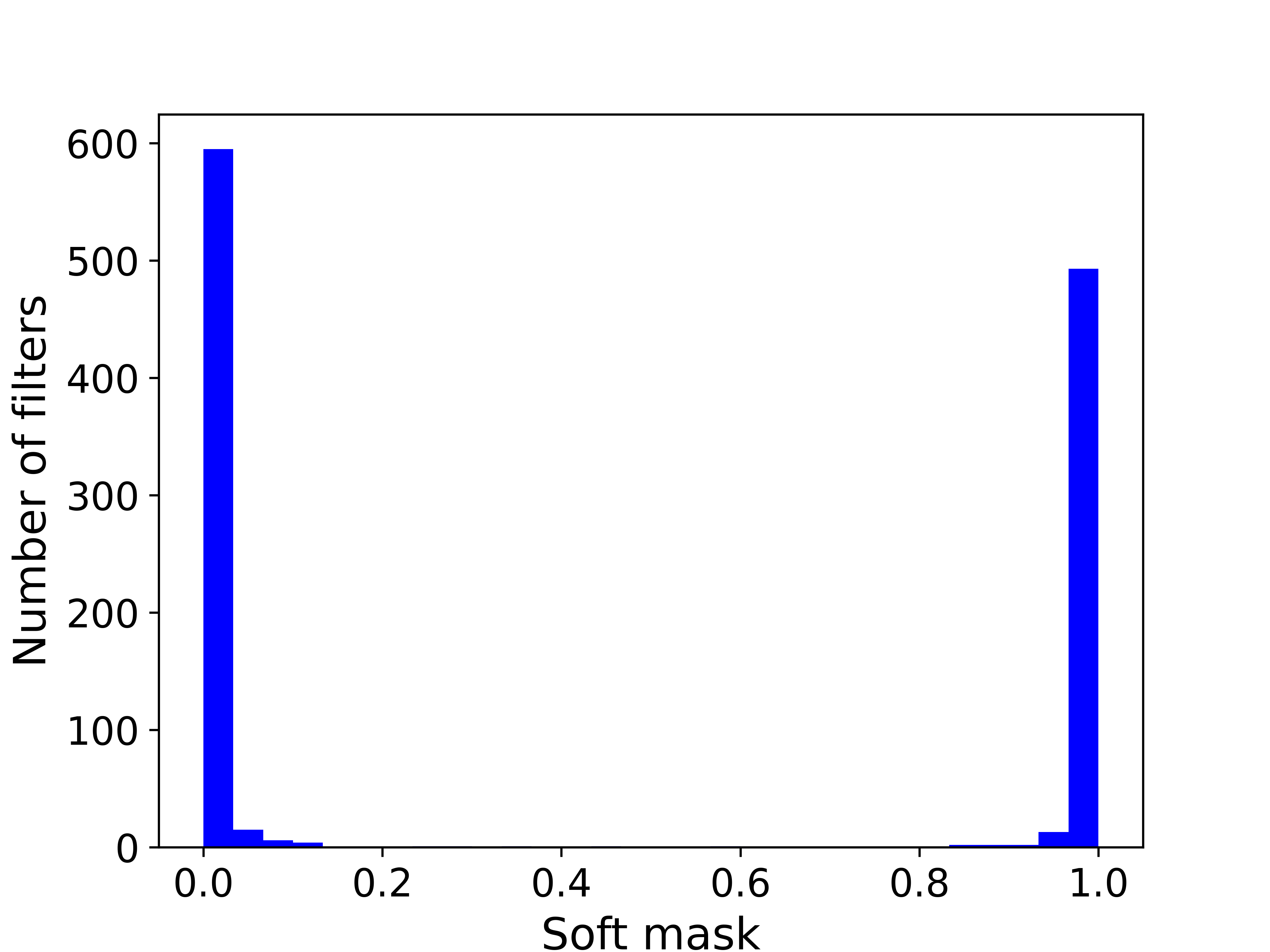}  
		\end{minipage}
	}
	\subfigure
	{
		\begin{minipage}{3.7cm}
			\centering      
			\includegraphics[scale=0.032]{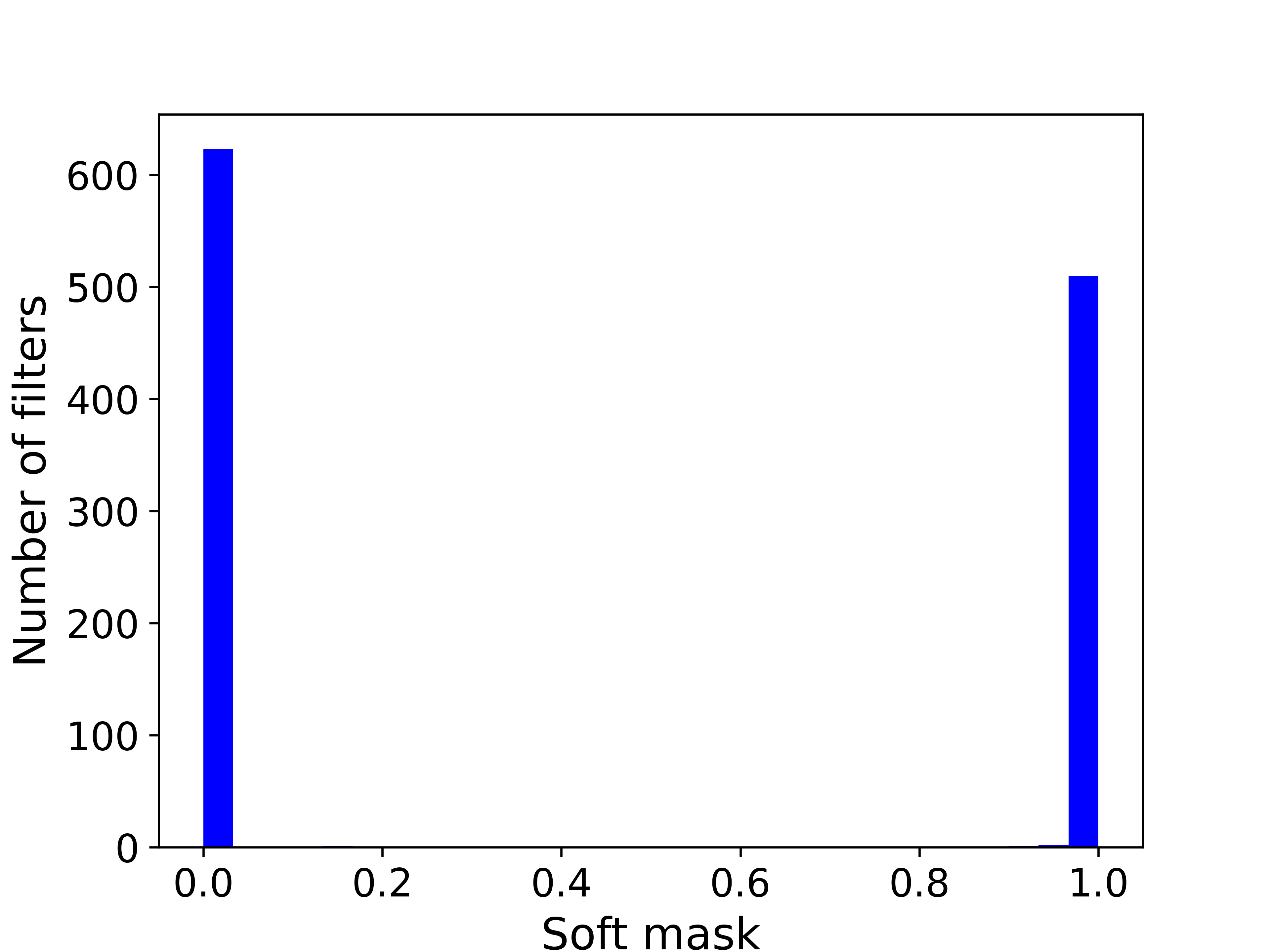}  
		\end{minipage}
	}
	\caption{The polarization effect of the proposed regularizer on the soft masks. Histograms of the soft masks are plotted at different iterations during the training process. With the training iterations increase, the soft masks are gradually separated into two parts with a large margin between them, and finally the masks of some filters locate at near 0 and the remaining locate at 1.} 
	\label{fig6}
\end{figure*}

\begin{figure*}[tb]
	\centering
	\includegraphics[width=0.98\textwidth]{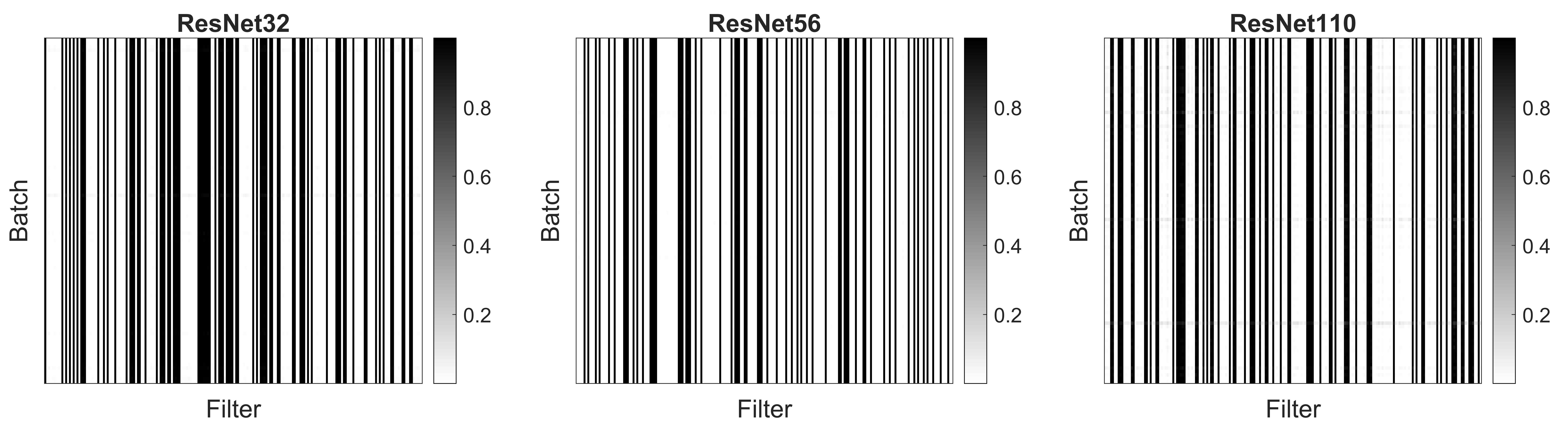}
	\caption{Analysis of similarity between the soft masks across different batches. 100 batches of inputs and 200 filters are randomly selected for evaluation, each row represents a batch and each column denotes a filter. Values of the soft mask derived from different batches are compared for each filter on CIFAR-10 dataset.}
	\label{fig7}
\end{figure*}

\textbf{The effectiveness of the regularizer on the soft masks.} To verify the effectiveness of our proposed regularizer on the soft masks of the filters, we conduct several experiments on CIFAR-10 dataset to compare our pruning results of ResNet32 and ResNet56 networks with the results obtained by replacing our proposed regularizer with other regularizers. Specifically, $\ell_1$ regularizer and the polarization regularizer proposed in~\cite{zhuang2020neuron} (here named as Zhuang polarization) are adopted for evaluation. For the pruning results based on the $\ell_1$ regularizer, there is no obvious gap in the mask distribution (as shown in Figure~\ref{fig5}), therefore we manually select the threshold on the masks to achieve the desired FLOPs reduction rate. For the pruning results based on the Zhuang polarization, the filters are separated into two parts due to the effect of polarization (as shown in Figure~\ref{fig5}), and we follow the steps introduced in~\cite{zhuang2020neuron} to select the threshold. As shown in Table~\ref{tab5}, the proposed polarization regularizer yields better performing subnetworks that has higher accuracy and similar or higher FLOPs reduction rate. With $\sim$63\% FLOPs reduction rate on ResNet32, applying our regularizer enables 0.78\% accuracy boost, while the accuracy improvement is 0.07\% for the $\ell_1$ regularizer and 0.24\% for the Zhuang polarization. Our method also yields compact subnetworks for ResNet56 that show higher accuracy than those generated by the competitors. For instance, with $\sim$66\% FLOPs removed, our polarization regularizer improves the model accuracy by 0.33\%, while the $\ell_1$ regularizer results in accuracy loss and the Zhuang polarization improves the accuracy by 0.18\% with a lower FLOPs reduction rate. Compared to the polarization results obtained by the Zhuang polarization, our method makes the filters be gradually separated into two parts when the training iterations increases, and finally a large margin between the two sets of filters is observed (as shown in Figure~\ref{fig6}). As our proposed polarization regularizer encourages the masks to be located at the two ends, the discretization gap between training and evaluation is significantly reduced, which makes our method get better performance than the Zhuang polarization. These results demonstrate the proposed regularizer is effective in identifying important filters, and convenient to automatically remove redundant filters without manual exploring of the threshold.

As our method uses the weighted average of soft masks calculated for each batch of inputs to scale channel-wise outputs of the pruned network, and different batches may end up with differently distributed soft masks, we further evaluate the similarity between the soft masks of different batches after the model converges. Specifically, the evaluation is performed on CIFAR-10 dataset, and we randomly selected 100 batches and 200 filters for comparison. The results on Figure~\ref{fig7} show different batches yield highly consistent distribution of the soft masks, and the filters are consistently separated into two parts with a large margin across different batches, which suggests our method has high robustness in reasoning the redundant filers and only one batch of inputs is required to prune the filters after model converges.

\textbf{Measuring filter importance with only simple or complex instances.} 
We further make a comparison between different pruned models with only simple or complex instances used for measuring the filter importance. As described in section 3.2, the complexity of each instance is calculated as the CE loss, and instances with high loss are considered as complex instances. Specifically, the mean of the complexity is used as the threshold to empirically distinguish simple instances from complex ones. We then conduct two experiments to compare the effects of simple and complex instances on the filter importance. In one of the experiments we set the weights of instances above the threshold to a small value of 1 $\times$ $10^{-5}$, and in another experiment we set the weights of instances below the threshold to 1 $\times$ $10^{-5}$, which implicitly eliminates either complex or simple instances when calculating the soft masks.

As shown in Table~\ref{tab6}, when silencing complex instances with a small weight, the accuracy is only 66.35\% without fine-tuning after pruning $\sim$50\% FLOPs, while the accuracy can be increased to 74.56\% at similar FLOPs reduction rate when only simple instances are suppressed. After fine-tuning of the pruned networks, the accuracy of the former is still 0.79 lower than the latter. These results demonstrate hard instances are more important to minimize the performance drop of network pruning.

\section{Conclusions}
In this paper, we propose a novel channel pruning method called CWP that uses instance complexity weighted soft masks to compress a large baseline network. The main novelty of CWP lies in its unique feature of modeling non-uniform contribution of different instances to the filter importance. By introducing instance-complexity related weights, the importance scores of filters are defined as the weighted sum of instance-specific soft masks, and hard instances are given higher weights in determining which filters should be preserved. In addition, a new regularizer is proposed in CWP to push the soft masks towards either 0 or 1, such that the sweet spot of 0.5 can be used to divide the filters into important and redundant filters with no need to manually search a threshold. Performance evaluations on various network architectures and datasets demonstrate CWP has advantages over the SOTAs in preserving model accuracy and pruning FLOPs. 

One of the limitations of CWP lies in the fact that it still requires a pretrained baseline network to calculate the weights of instances based on the CE loss, and pretraining a large baseline network often requires extensive computations. There may be other more efficient strategies to measure the weights of instances, such as dynamically adjust instance weights based on the network weights learned in last epoch, and we plan to investigate this in near future. In addition, we currently test our method on some common datasets, and we plan to apply our method to more datasets to examine its effectiveness on other research domains. Another limitation of CWP is that the hyper-parameters related to FLOPs reduction rate need to be manually selected, which may cause inconvenience when pruning large baseline networks, and we plan to introduce a procedure for automatically selecting the hyper-parameters for the desired FLOPs pruning rate in future versions of CWP.

\section*{Declaration of Competing Interest}
The authors declare that they have no known competing financial interests or personal relationships that could have appeared to influence the work reported in this paper.

\section*{Acknowledgements}
This work has been supported in part by the National Natural Science Foundation of China (61901238), West Light Foundation of The Chinese Academy of Sciences (XAB2019AW12), and Key Research and Development Program of Ningxia (2021BEE03013).





		

%






\bibliographystyle{apalike}
\bibliography{reference}




\end{document}